\pdfoutput=1

\documentclass[11pt]{article}

\usepackage[]{acl}

\usepackage{times}
\usepackage{latexsym}
\usepackage{float}

\usepackage[T1]{fontenc}

\usepackage[utf8]{inputenc}

\usepackage{microtype}

\usepackage{inconsolata}

\usepackage{booktabs}
\usepackage{graphicx}
\usepackage{longtable}
\usepackage{tcolorbox}
\usepackage{stackengine}
\usepackage{multirow}
\usepackage{graphicx}
\usepackage{wrapfig}
\usepackage{hyperref}
\usepackage{makecell}
\usepackage{booktabs} 
\usepackage{array}    
\usepackage{geometry} 

\usepackage{url}            
\usepackage{amsfonts}       
\usepackage{nicefrac}       
\usepackage{microtype}      
\usepackage{xcolor,colortbl}         
\usepackage{times}
\usepackage{latexsym}
\usepackage{multicol}
\usepackage{blindtext}
\usepackage{tabu}
\usepackage{amsmath, bm}

\usepackage{subcaption}
\usepackage{caption}
\usepackage[normalem]{ulem}
\usepackage{soul}
\usepackage[shortlabels]{enumitem}
\usepackage{array}
\usepackage{pgffor}
\usepackage{textcomp}
\usepackage{amssymb}
\usepackage{pifont}
\usepackage{booktabs}
\usepackage{tabularx}

\definecolor{lightgreen}{rgb}{0.8, 0.95, 0.8}
\definecolor{lightred}{rgb}{0.95, 0.8, 0.8}
\definecolor{naplesyellow}{rgb}{0.98, 0.85, 0.37}
\definecolor{pastelyellow}{rgb}{0.99, 0.99, 0.59}

\title{Step-by-Step Reasoning to Solve Grid Puzzles: Where do LLMs Falter?}

\author{Nemika Tyagi$^{1*}$ \quad Mihir Parmar$^{1*}$ \quad Mohith Kulkarni$^1$ \quad Aswin RRV$^1$ \quad Nisarg Patel$^1$ \\ \textbf{Mutsumi Nakamura}$^1$ \quad \textbf{Arindam Mitra}$^2$ \quad \textbf{Chitta Baral}$^1$ \\\\ 
$^1$Arizona State University \quad $^2$Microsoft Research\\
\small{\texttt{\{ntyagi8, mparmar3, chitta\}@asu.edu}}
}

\begin{document}
\maketitle
\begin{abstract}
Solving grid puzzles involves a significant amount of logical reasoning. Hence, it is a good domain to evaluate the reasoning capability of a model which can then guide us to improve the reasoning ability of models. However, most existing works evaluate only the final predicted answer of a puzzle, without delving into an in-depth analysis of the LLMs' reasoning chains (such as where they falter) or providing any finer metrics to evaluate them. Since LLMs may rely on simple heuristics or artifacts to predict the final answer, it is crucial to evaluate the generated reasoning chain beyond overall correctness measures, for accurately evaluating the reasoning abilities of LLMs. To this end, we first develop \textit{GridPuzzle}, an evaluation dataset comprising 274 grid-based puzzles with different complexities. Second, we propose a new error taxonomy derived from manual analysis of reasoning chains from LLMs including GPT-4, Claude-3, Gemini, Mistral, and Llama-2. Then, we develop an LLM-based framework for large-scale subjective evaluation (i.e., identifying errors) and an objective metric, \textit{PuzzleEval}, to evaluate the correctness of reasoning chains. Evaluating reasoning chains from LLMs leads to several interesting findings. We further show that existing prompting methods used for enhancing models' reasoning abilities do not improve performance on \textit{GridPuzzle}. This highlights the importance of understanding fine-grained errors and presents a challenge for future research to enhance LLMs' puzzle-solving abilities by developing methods that address these errors\footnote{Data and source code are available at \url{https://github.com/Mihir3009/GridPuzzle}}.

\def\thefootnote{*}\footnotetext{Equal Contribution}\def\thefootnote{\english{footnote}}

\end{abstract}

\section{Introduction}
\label{sec:intro}

Recent advancements in LLMs such as GPT-4, Gemini, Claude-3 \cite{anthropic2024claude}, Llama-2 \cite{touvron2023llama}, and Mistral \cite{jiang2023mistral} have achieved remarkable performance on a wide range of Natural Language Understanding (NLU) tasks previously thought to be exclusive to humans. Beyond NLU, exploring LLMs' logical reasoning abilities \cite{liu2021logiqa, saparov2022language, parmar2024towards, patel2024multi} on complex tasks such as puzzle-solving is under-explored. Past attempts have been made to evaluate models on logic-intensive grid-based puzzle-solving. However, they either do not focus on evaluating LLMs \cite{mitra-baral-2015-learning, jabrayilzade-tekir-2020-lgpsolver} or do not evaluate LLMs independently, but rather use neuro-symbolic approaches \cite{ishay2023leveraging} that use external specialized solvers on LLM outputs. Here, we aim to evaluate the puzzle-solving abilities of LLMs by themselves, without the use of any external logic solvers.

To understand the reasoning capabilities of LLMs, it is important to evaluate reasoning chains, rather than the final predicted answer. There have been works that evaluate reasoning chains using objective metrics such as ROSCOE \cite{golovneva2022roscoe}, CTC \cite{deng-etal-2021-compression}, and BARTScore \cite{yuan2021bartscore}, however, they do not focus specifically on evaluating reasoning. Some prior works propose metrics for specific reasoning tasks, such as FOLIO \cite{han2022folio} and ProntoQA \cite{saparov2022language}. However, these methods rely on reference-based evaluation, do not focus on puzzle-solving, and do not aim to identify fine-grained errors in reasoning chains. To address these limitations, we propose a reference-free manual and automated subjective evaluation of reasoning chains to understand various fine-grained errors in reasoning chains for grid-based puzzle-solving.

\begin{figure*}
    \centering
    \includegraphics[width=0.95\linewidth]{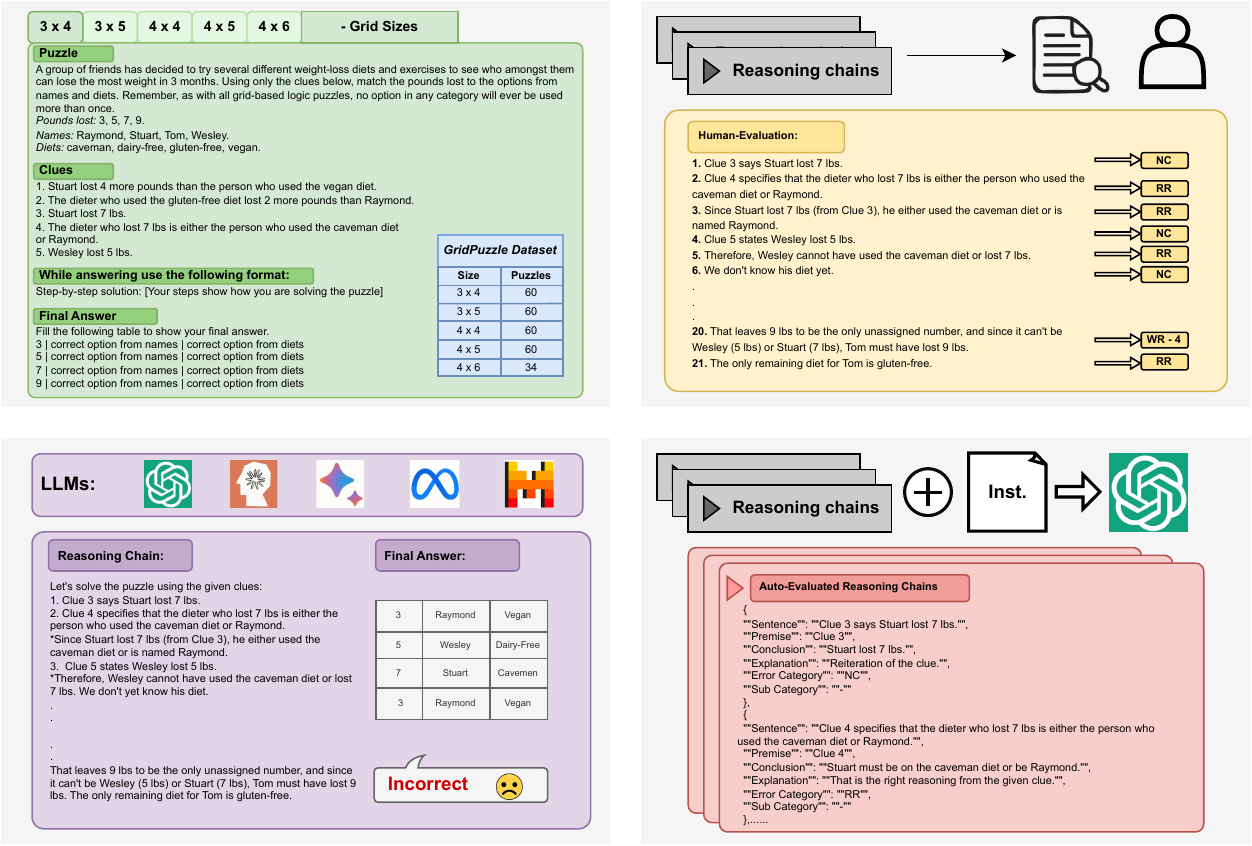}
    \caption{Schematic representation of proposed pipeline. Begins with the data collection of \textit{GridPuzzle} dataset (top left) and evaluating various LLMs in zero-shot CoT setting (bottom left), then analyzing reasoning chains of LLMs manually to find various error types (top right) and automate this analysis process using LLM to check the correctness of reasoning chain by finding errors (bottom right).}
    \label{fig:teaser}
\end{figure*}

Motivated by \citet{mitra-baral-2015-learning}, we first develop \textit{GridPuzzle} (Figure \ref{fig:teaser}), a comprehensive evaluation dataset consisting of grid-based puzzles with grid-size of $3 \times 4$, $3 \times 5$, $4 \times 4$, $4 \times 5$, and $4 \times 6$ with three levels of difficulty (easy, medium, and hard). Then, we evaluate LLMs including GPT-4, Gemini-Pro, Claude-3, Llama-2, and Mistral on \textit{GridPuzzle} in zero-shot-CoT setting (Figure \ref{fig:teaser}). Experimental results show that LLMs do not fare well and achieve a maximum of $5.1\%$ accuracy. 


To investigate the reasoning chains, we manually analyze them (Figure \ref{fig:teaser}) to find fine-grained errors (further details in section \ref{sec:errors_cat}). Based on this, we propose a new error taxonomy comprising five broad categories, and nine fine-grained sub-categories (Tables \ref{tab:broad_category} and \ref{tab:sub_category}), providing deeper insights into the primary causes of the LLMs' reasoning failures. However, scaling manual analysis to a larger set is time-consuming and laborious. Hence, we propose to leverage LLMs as auto-evaluators by creating prompts that utilize error taxonomy to automate the analysis of reasoning chains and help in identifying errors (Figure \ref{fig:teaser}). While evaluating w.r.t. manual annotation, our auto-evaluator model achieves $\sim86\%$ agreement, hence providing quality error categorization.


Beyond identifying errors and the accuracy of the final answer, we propose \textit{PuzzleEval}, an LLM-based framework to evaluate reasoning chains for grid-based puzzles. \textit{PuzzleEval} involves a multi-stage evaluation using GPT-4o. First, we identify key logical conclusions from the reasoning chain; second, we extract key logical concepts from these conclusions; and finally, we measure the presence of these logical concepts in the final gold answer to assess the correctness of the reasoning chain. Evaluating reasoning chains based on error categorization and \textit{PuzzleEval} reveals interesting findings such as LLMs show lower accuracy despite having more error-free reasoning steps, open-source models lack reasoning skills compared to closed-source models, and the most dominant error categories are wrong reasoning and elimination. Additionally, we employ existing prompting methods such as Plan-and-Solve and Self-discover, demonstrating that these methods do not improve performance on \textit{GridPuzzle}. We believe that our findings will inspire future work in the automated in-depth evaluation of reasoning chains for broader reasoning tasks and enhance the reasoning abilities of models.

\section{Related Work}
\label{sec:related_work}

\paragraph{Puzzle-solving Task}

Puzzle-solving task provides insights into LLMs’ logical reasoning. \citet{giadikiaroglou2024puzzle} categorize puzzles into (1) rule-based and (2) rule-less puzzles. Rule-less puzzles include riddles \cite{lin-etal-2021-riddlesense}, MCQs \cite{zhao2023solving}, programming puzzles \cite{schuster2021programming}, and commonsense reasoning puzzles \cite{gu2023beyond}; however, in our work we focus on rule-based puzzles. In rule-based puzzles, past attempts have explored Sudoku \cite{Noever2021PuzzleSW}, Rubik's Cube, 8-puzzle, Game of 24 \cite{yao2024tree}, crosswords \cite{yao2024tree}, chess puzzles \cite{feng2024chessgpt}, card games \cite{gupta2023chatgpt}, BoardgameQA \cite{kazemi2024boardgameqa}, and Lateral Thinking Puzzles \cite{huang2024latevalinteractivellmsevaluation}. However, grid-based puzzle solving is under-explored. \citet{mitra-baral-2015-learning} proposed a grid-based puzzle dataset and \citet{dziri2023faithfatelimitstransformers} studied compositionality in LLMs using Grid Puzzle, but these works do not provide any insights into the performance of recent LLMs. Motivated by this, we propose a systematically curated grid-based puzzle dataset, \textit{GridPuzzle}, and provide an evaluation of various LLMs in puzzle-solving.

\paragraph{Automatic Evaluation of Reasoning Chains}

Previous works \cite{dalvi-etal-2021-explaining, saparov2022language, han2022folio} have focused on reference-free evaluation, which is not reliant on gold-reasoning chains. Recently, ROSCOE \cite{golovneva2022roscoe} proposed a suite of metrics to measure the semantic consistency, logicality, informativeness, fluency, and factuality of reasoning chains, while the ReCEval framework \cite{prasad-etal-2023-receval} evaluates reasoning chains based on two key properties: correctness and informativeness. Recent evaluation methods such as \textit{LLM evaluation} \cite{chiang-lee-2023-large} and G-Eval \cite{liu-etal-2023-g} leverage LLMs to measure the quality of reasoning chains.  \textit{LLM evaluation} involves presenting task instructions and a text sample to LLMs, asking them to rate the sample's quality on a 5-point Likert scale, whereas the latter incorporates automatic chain-of-thought generated by the LLM describing the detailed evaluation steps. Additionally, \citet{tyen2023llms}’s attempt to use GPT-4 as an evaluator in a few-shot setting, shows that evaluating reasoning chains remains a challenge.  Furthermore, AutoRace (Automatic Reasoning Chain Evaluation) \cite{hao2024llm} proposed a fully automated approach for evaluating reasoning chains that adapt to different tasks without human effort. However, these methods do not evaluate reasoning chains at the level of fine-grained error types and do not provide detailed task-specific insights. To address this, we propose LLM-based reference-free evaluation methods that identify fine-grained errors and assess the correctness of generated reasoning chains.

\section{Evaluation of Reasoning Chains}
\label{sec:error_category}

\subsection{\textit{GridPuzzle}}

To develop this dataset, we extract logic grid puzzles of various grid sizes from Puzzle Baron's Logic Puzzles\footnote{\url{https://logic.puzzlebaron.com/}}. Specifically, we compile logic grid puzzles of size $3 \times 4$, $3 \times 5$, $4 \times 4$, $4 \times 5$, and $4 \times 6$. Each grid size has three levels of difficulty (easy, medium, and hard) except $4 \times 6$. This particular grid size has only two difficulty levels (Easy and Medium). Statistics corresponding to each grid size are presented in Figure \ref{fig:teaser} (top left).
\begin{table}[!htbp]
\centering
\resizebox{\linewidth}{!}{
\begin{tabular}{c|c}
\toprule
\begin{tabular}[c]{@{}c@{}}\textbf{Error}\\ \textbf{Category}\end{tabular}
& \textbf{Description} \\
\midrule
\textbf{WW} & Wrong Premise and Wrong Conclusion \\
\textbf{WR} & Wrong Premise and Right Conclusion \\
\textbf{RW} & Right Premise and Wrong Conclusion \\
\textbf{RR} & Right Premise and Right Conclusion \\
\textbf{NC} & No Conclusion statement or no reasoning involved \\
\bottomrule
\end{tabular}
}
\caption{Proposed error taxonomy for broad categories based on manual analysis. If a statement starts with ``so, therefore, hence, this means, this implies, etc.'' and/or is not followed by any premise, consider the previous statement’s conclusion or the previous NC as the premise.}
\label{tab:broad_category}
\end{table}



\begin{table*}
    \centering
    \resizebox{\linewidth}{!}{%
    \begin{tabular}{c|c|c|>{\centering\arraybackslash}m{11cm}}
        \toprule
        \textbf{Category} & \textbf{Source} & \textbf{Sub-Category} & \textbf{Description} \\ \midrule\midrule
        \multirow{9}{*}{\begin{tabular}[c]{@{}c@{}}Wrong Premise or\\ No Conclusion\end{tabular}}
        & 
        \multirow{6}{*}{\begin{tabular}[c]{@{}c@{}}From the clues \\ (Example: From clue 4,....)\end{tabular}} & (1) Hallucination & When information is completely out of context and not present in clues. \\ \cmidrule{3-4}
         &  & (2) Incomplete Information & Lacks necessary information to make a particular conclusion. 
         \\ \cmidrule{3-4}
         &  & (3) Assumptions & Statements not derived from clues directly; might include assumed information relevant to the clue. \\ \cmidrule{2-4}
         & 
         \multirow{3}{*}{\begin{tabular}[c]{@{}c@{}}Derived Conclusions using \\clues given in puzzle \\which was not inherently\\ given in the clues.\end{tabular}}
         & (4) Error Propagation & Premise derived from a previous incorrect conclusion. \\ \cmidrule{3-4}
         &  & (5) Incomplete Information & Lacks necessary information to make a particular conclusion. \\ \cmidrule{3-4}
         &  & (6) Wrong Assumption & The derived assumption is incorrect. \\ \midrule
        \multirow{3}{*}{Wrong Conclusion} & 
        \multirow{3}{*}{\begin{tabular}[c]{@{}c@{}}Derived using the premise\\ (which itself is either\\ taken directly from the\\ clues or derived)\end{tabular}}
        & (a) Wrong Reasoning & The reasoning is incorrect, regardless of the premise’s accuracy. \\ \cmidrule{3-4}
         &  & (b) Error propagation & Conclusion is incorrect due to an erroneous premise. \\ \cmidrule{3-4}
         &  & (c) Wrong Elimination & All premises are present, but not all conclusions are correctly derived. \\ \bottomrule
    \end{tabular}
    }
    \caption{Proposed error taxonomy for sub-categories based on manual analysis. These sub-categories are defined for cases where either the conclusion or premise is incorrect (``RW'' or ``WR'') or both are incorrect (``WW''). For ``WW'', the error sub-categories might appear in any combinations between (1-6) and (a-c) such as `1a', `4b', or `6c'.}
    \label{tab:sub_category}
\end{table*}
\subsection{Manual Evaluation}

To explore where exactly these LLMs falter in performing reasoning, we conduct a detailed manual analysis of the reasoning chains generated by them while solving grid-based puzzles. Details of the annotation guidelines provided to the human evaluators are given in the Appendix \ref{Appendix:guideline}. Our manual analysis process consists of three steps. First, we begin by segmenting the reasoning chains into individual sentences, allowing us to categorize errors more precisely. Second, we identify the premise and conclusion for each sentence and determine their respective correctness. We refrain from subdividing sentences into multiple premises or conclusions to maintain simplicity for finding errors. At last, each sentence is categorized as either containing a single premise and conclusion or being a declarative statement without a conclusion. Then, we begin assessing potential issues or errors in the reasoning chains. Now, we follow an exhaustive approach to create fine-grained error categories. We begin with 30 reasoning chains (6 puzzles x 5 reasoning chains from LLMs) to manually identify potential errors. Next, we categorize these errors in a structured format. We then add another 30 reasoning chains to see if any new types of errors emerge. If new errors are identified, we refine our categories accordingly. This process is repeated until we evaluate a total of 150 reasoning chains and no new types of errors are found. Based on this method, we have carefully filtered and categorized several errors made by LLMs, presenting them as five broad categories and nine sub-categories.

\subsection{Proposed Error Taxonomy}
\label{sec:errors_cat}


\paragraph{Broad Categories} As shown in Table \ref{tab:broad_category}, we present five main categories: ``WW'' - \textit{Wrong Premise Wrong Conclusion}, ``WR'' - \textit{Wrong Premise Right Conclusion}, ``RW'' - \textit{Right Premise Wrong Conclusion}, ``RR'' - \textit{Right Premise Right Conclusion}, and ``NC'' - \textit{No Conclusion}. These acronyms of broad categories are self-explanatory. For instance, the category ``WW'' comprises sentences where the sentence consists of a wrong premise as well as a wrong conclusion. Interestingly, we also find the ``WR'' category consists of instances where a wrong premise still leads to a correct conclusion. Additionally, sentences containing only information from clues or premises from previous steps fall under ``NC''. We conduct further investigation as to why the premises and conclusions become incorrect.



\paragraph{Sub-categories: Wrong Premise}  As shown in Table \ref{tab:sub_category}, we identified the source of the premise to determine the origin of errors: (i) ‘\textbf{\textit{From Clues}}’ – where the premise is directly borrowed from one of the clues without any further reasoning, and (ii) ‘\textbf{\textit{Derived}}’ – where the premise is inferred from either the clues or the previous conclusions. From Table \ref{tab:sub_category}, there are six possible reasons associated with two different sources for the wrong premise. When the premise originates from the source (i), we find three types of errors: \textbf{Hallucination} – When some factual information from the clues is distorted or completely made up; \textbf{Incomplete information} – When the information is correctly borrowed from the clues but it is not sufficient to make a particular conclusion; \textbf{Assumptions} – This is a special category where the premise is not derived but also not given exactly in the clues. It is often related to one of the clues and is of the form, “Let’s assume” or “Assuming that.” When source is \textit{derived}, we find three different errors: \textbf{Error Propagation} – This occurs when a previously incorrect conclusion becomes the basis for a flawed premise, thereby extending the error from one conclusion to the next; \textbf{Incomplete information} – When the derived premise is not sufficient to make a particular conclusion; and lastly, \textbf{Wrong Assumption} – When the LLM reasoner clearly states that a premise was an assumption but it was incorrectly derived. 

\paragraph{Sub-categories: Wrong Conclusion} As shown in Table \ref{tab:sub_category} (source), conclusions are always logically derived from a fixed set of premises. For having a wrong conclusion in any reasoning step, we find 3 errors responsible: \textbf{Error Propagation} – When a conclusion is wrong strictly due to some error in the preceding premise; \textbf{Wrong Elimination} – When the conclusion is wrong because the LLM reasoner failed to eliminate all the unfit choices correctly. This case is specific to the grid-based puzzle task but is inherently an erroneous deduction on the LLM’s end; \textbf{Wrong reasoning} – The remaining incorrect conclusions that did not fit in the above categories are classified under this label.


\begin{table}
\small
\resizebox{0.9\linewidth}{!}{
\begin{tabular}{p{0.85\columnwidth} }
\toprule
    \textbf{Examples of reasoning chain evaluated by GPT-4o} \\ \midrule


       \multirow{6}{*}{}{\textbf{Sentence:}"Therefore, Zeno must be 69\%, and UCLA must be 62\%.",}\\ \textbf{Premise:} "If Zeno were 55\%, there would be no score 7\% lower than 55\% for UCLA.", \\ \textbf{Conclusion:} "Zeno must be 69\%, and UCLA must be 62\%", \\ \textbf{Explanation:} "The conclusion is incorrect as UCLA is already known to be 62\% from clue 2.", \\ \textbf{Error Category:} "RW.", \\ \textbf{Premise:} "A", \\ 
    \midrule

        \multirow{6}{*}{}{\textbf{Sentence:}"Since the third performer used flashlights, it must be either Lora or Carmen.",}\\ \textbf{Premise:} "The performer who used flashlights was either Lora or Carmen.", \\ \textbf{Conclusion:} "The third performer must be either Lora or Carmen.", \\ \textbf{Explanation:} "The conclusion is based on the incorrect premise that the third performer used flashlights.", \\ \textbf{Error Category:} "WW", \\ \textbf{Sub Category} "4B",\\

    \bottomrule

\end{tabular}
}
\caption{Examples of reasoning chain evaluated by Auto-evaluator (GPT-4o).}
\label{tab:example_annot}
\end{table}

\subsection{Automated Evaluation}
Manual analysis of reasoning chains provides a detailed categorization of errors; however, it is tedious and, therefore, challenging to scale for the entire dataset. However, analyzing the distribution of errors from our proposed taxonomy on the whole dataset is also crucial in understanding the shortcomings of LLMs' reasoning ability. Thus we develop an LLM-based auto-evaluator to automate the process of error evaluation. To this end, we prompt the GPT-4o model to identify and categorize errors in the given reasoning chain. Our prompt consists of a system instruction followed by a user prompt containing the reasoning chain to be evaluated along with the original puzzle and its gold solution. The system prompt can be further dissected into 3 key components: the instructions, the knowledge, and an exemplar. The \textbf{instruction} contains all the rules that the GPT-4o needs to follow to conduct accurate evaluation and error categorization of the reasoning chains. It incorporates similar sequential steps used during the manual evaluation of reasoning chains along with the required output format. The \textbf{knowledge} has a detailed description of our error taxonomy including the broad and sub-categories. We also provide a preference order for selecting categories along with the description to minimize any ambiguity in the evaluation process. Lastly, the \textbf{exemplar} consists of a puzzle, its correct solution, the original model-produced reasoning chain, and the manually evaluated reasoning chain with our error categories. We termed this LLM-based evaluator as ``Auto-evaluator''. Appendix \ref{appendix:autoevaluator} provides the structure of the Auto-evaluator prompt. 


Using the Auto-evaluator, we evaluated a total of 1,370 reasoning chains generated by five different LLMs for solving 274 puzzles. The application of our Auto-evaluator to this large dataset allowed us to analyze the distribution of error categories on a broader scale. To validate the accuracy of the evaluations performed by the Auto-evaluator, we randomly sampled 20 reasoning chains from the manually evaluated set. The authors then compared their error category assignments to those given by the Auto-evaluator. The agreement score for the total number of reasoning steps between the manual evaluation and the GPT-4o evaluation is $\sim86\%$. Table \ref{tab:example_annot} shows the example of reasoning steps evaluated by GPT-4o.

\section{Experimental Setup}
\label{sec:exp}

\subsection{Experiments}
We evaluate a range of closed-source LLMs including GPT-4-Turbo, Claude-3-Opus, and Gemini-Pro, and open-source models Llama-2-13B, and Mistral-7B-Instruct on \textit{GridPuzzle} in the Zero-shot-CoT setting \cite{kojima2022large}. We also conducted a scaling experiment on Llama-2-70B and the results are given in the Appendix \ref{Appendix:llama70b}. Our \textit{GridPuzzle} dataset consists of a set of instances denoted as $\mathcal{P} = {<p_{n}^{i \times j}, a_n>}$, where $p_n^{i \times j}$ is $n^{th}$ puzzle instance with grid size of $i \times j$ and $a_n$ as a gold answer. We prompt each LLM to generate a reasoning chain before predicting answer $\hat{a}$. To evaluate each model in the Zero-shot-CoT setting, we provide $< I, p_n^{i \times j} >$ as input to the model and predict an answer $\hat{a}$ where $I$ is a natural language instruction. The evaluation is conducted on the OpenAI, Google, and Anthropic model versions released in April 2024 with temperature setting 0 for deterministic predictions. NVIDIA A100 GPUs are used for conducting the inference of open-source models with a batch size of 4. The example prompts used for these experiments are provided in Appendix \ref{appendix:puzzlesample}.
\begin{figure*}[!htbp]
    \centering
    \includegraphics[width=0.96\linewidth]{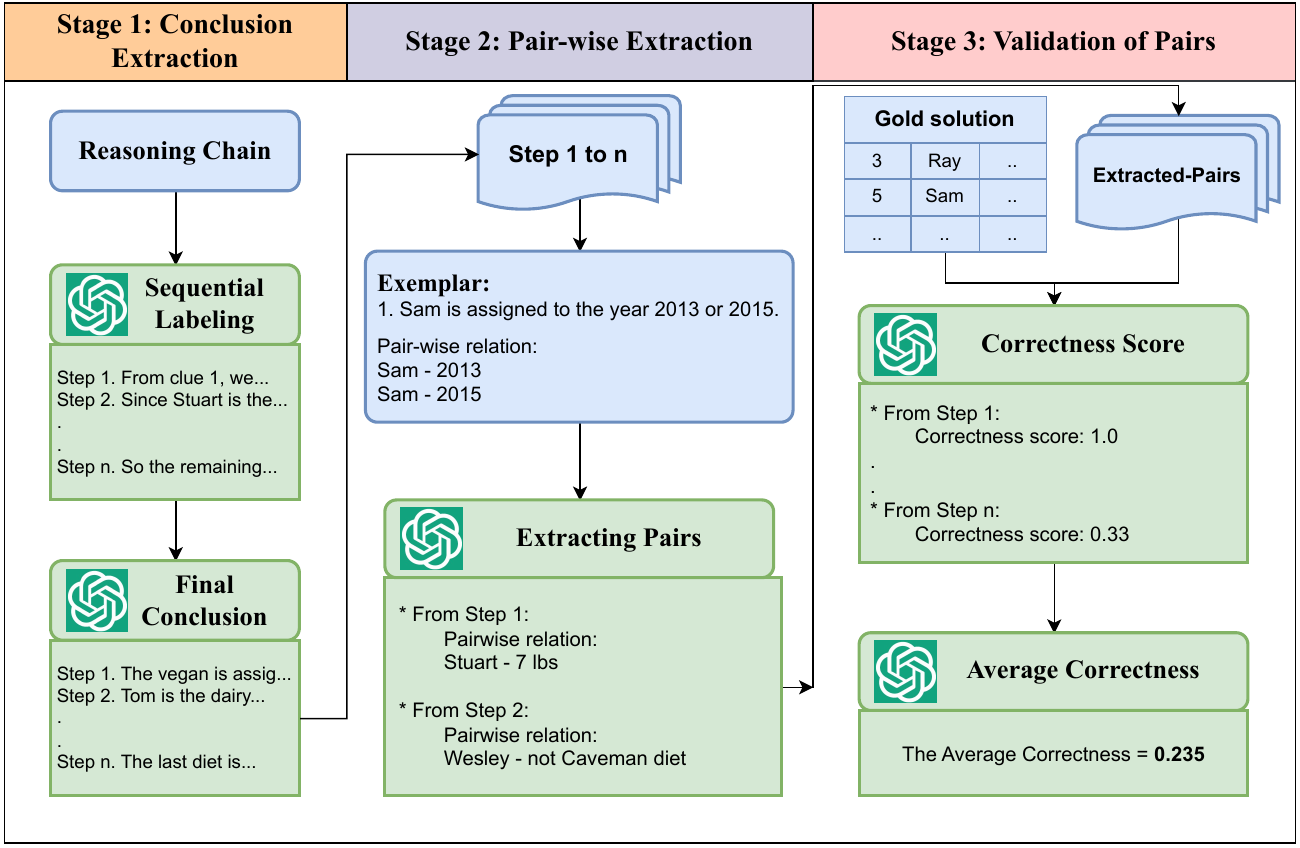}
    \caption{The process of calculating \textit{PuzzleEval} metrics is described above. The reasoning chains are produced by our five LLMs and the gold solution is taken from our \textit{GridPuzzle} dataset.}
    \label{fig:pemetrics}
\end{figure*}

\subsection{Metrics}

\paragraph{Accuracy} We use accuracy to demonstrate the capability of LLMs in solving grid-based puzzles based on their ability to predict the final answer. To calculate this metric, we use the LLM-generated final answers and compare them with the available gold solution. The predicted answers and the gold solution are in the form of tables with the number of rows and columns equal to the grid size of the puzzle. We perform an Exact Match (EM) to compare the two tables and mark them as correct only when all the entries of the tables match. See the example of the final answer table in Appendix \ref{Appendix:rcevaluation}. 

\begin{figure}
    \centering
    \includegraphics[width=0.8\linewidth]{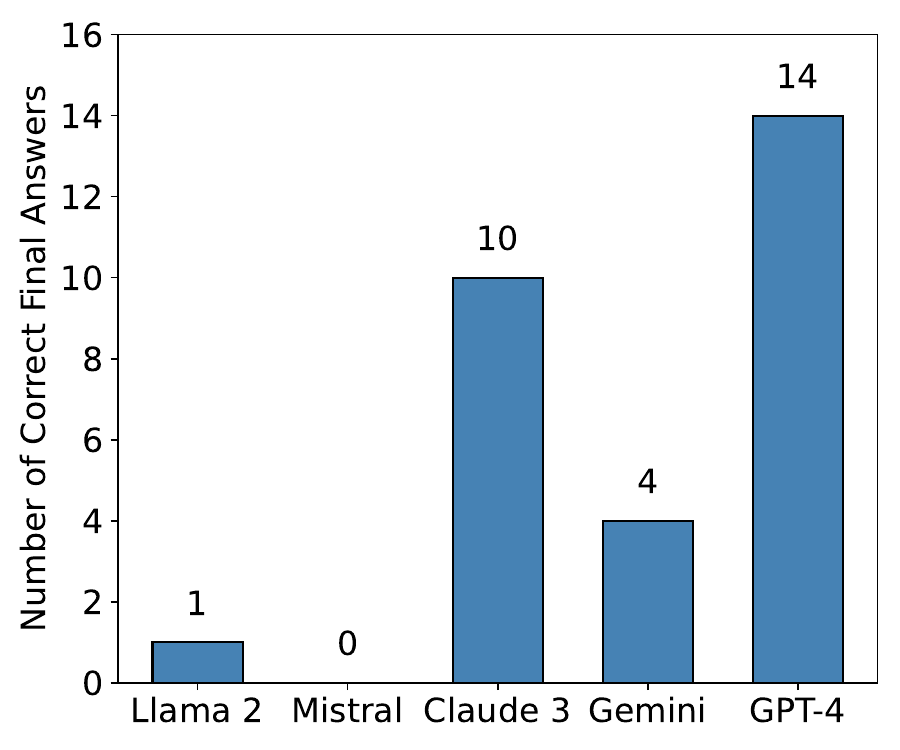}
    \caption{Performance of five different LLMs in terms of accuracy on the \textit{GridPuzzle} dataset.}
    \label{fig:mainresult}
\end{figure}
\label{sec:results}

\paragraph{\textit{PuzzleEval}}
We developed this LLM-based metric to assess step-by-step reasoning chains and provide a \textit{correctness score} for each step, as well as the Average Correctness Score (ACS) for the entire chain. \textit{PuzzleEval} is a reference-free metric specifically designed for assessing reasoning chains generated for grid-based puzzle tasks. It evaluates the correctness of each step in the reasoning chain and reports the score using only the final answer table provided as the gold solution, without requiring any comparison to a gold reasoning chain.


As shown in Figure \ref{fig:pemetrics}, \textit{PuzzleEval} consists of a three-stage pipeline to evaluate any reasoning chain. First, we prompt GPT-4o to label all the steps sequentially to account for any discrepancies in the different formats of reasoning chains produced by various models, and to extract only the final conclusions from each step. This stage is crucial as it removes the portion of a step where the models just reiterated clues or previous conclusions. Second, we instruct the model to extract the pair-wise relation of elements from the puzzle that have been either accepted or rejected in the extracted final conclusions. If the extracted conclusion is \textit{"Sam is assigned to the year 2015 but not 2014."}, these pairs are of the form “Sam – 2015” or “Sam – not 2014”. Third, we provide the gold solution table and ask the model to check if these accepted or rejected pairs match the given information. As per the validation, the pairs extracted from every step are marked as correct or incorrect. After obtaining this information for each step the \textit{correctness score} is calculated by adding up all the correct and incorrect steps (correct pairs are marked 1 and incorrect pairs are marked 0) divided by the total number of pairs in each step. Finally, the ACS is determined by adding up all the \textit{correctness scores} from each step and dividing by the number of steps to capture the overall quality of the reasoning chain. Hence, \textit{PuzzleEval} provides ACS for each reasoning chain in range of 0 to 1. 

\section{Results and Analysis}

\subsection{Objective Evaluation}
To evaluate the performance of LLMs when solving grid-based puzzles, we assess the outputs of 5 LLMs using the accuracy and \textit{PuzzleEval}. As shown in Figure \ref{fig:mainresult}, we found that all the models have low performance on the \textit{GridPuzzle} dataset in terms of accuracy. The smaller open-source LLMs completely failed at the puzzle-solving task, with Llama-2 solving only one puzzle correctly. Close-source models with significantly larger parameter sizes also exhibited poor performance. GPT-4 had the highest accuracy at only 5.11\% (14 puzzles out of 274). Despite the overall low performance of all LLMs, the closed-source models perform marginally better. We evaluate the quality of the reasoning chains using \textit{PuzzleEval}. Table \ref{tab:puzzleeval} provides the ACS for each grid size available in the \textit{GridPuzzle}. Surprisingly, compared to the accuracy, the performance of the models with \textit{PuzzleEval} was significantly better as shown in Table \ref{tab:puzzleeval}. The ACS lie in the range of 0.26 to 0.64 across all grid sizes. This higher score can be attributed to the partial correctness of reasoning chains when solving the grid-puzzle task. The disparity between metrics shows that evaluating only final answers doesn't fully capture LLMs' effectiveness in complex logical tasks like grid puzzles.

\begin{table}
\small
\begin{center}
\begin{tabular}{c|ccccc|c}
\toprule
\textbf{Model}   & \textbf{3 x 4} & \textbf{3 x 5} & \textbf{4 x 4} & \textbf{4 x 5} & \textbf{4 x 6}  & \textbf{Avg} \\ 
\midrule
Llama   & 0.45  & 0.46  & 0.46  & 0.42  & 0.28  & \textbf{0.41} \\ 
Mistral & 0.29  & 0.26  & 0.27  & 0.26  & 0.27  & \textbf{0.27} \\ 
Claude  & 0.60  & 0.56  & 0.52  & 0.55  & 0.46  & \textbf{0.54} \\ 
Gemini  & 0.60  & 0.64  & 0.54  & 0.52  & 0.62  & \textbf{0.58} \\ 
GPT-4   & 0.61  & 0.62  & 0.56  & 0.54  & 0.60  & \textbf{0.59} \\ 

\bottomrule
\end{tabular}
\end{center}
\caption{The results for \textit{PuzzleEval} on the different grid sizes available in \textit{GridPuzzle} dataset in terms of ACS.}
\label{tab:puzzleeval}
\end{table}

With the increase in the sizes of the grids, the complexity of the puzzles also rises, leading to a depreciating performance by the LLMs with larger grids. Overall the performance of larger LLMs was much better than the small open-source models. Mistral-7B performed the worst in \textit{PuzzleEval} which is in accordance with its low accuracy score. GPT-4 and Gemini models surprisingly have similar \textit{PuzzleEval} scores (0.59 and 0.58 respectively) despite their large difference in accuracy. This difference in \textit{PuzzleEval} could be attributed to the relatively shorter reasoning chains (fewer reasoning steps) produced by Gemini (an average of $14.91$ steps) compared to GPT-4 (an average of $20.66$ steps). Shorter reasoning chains may reduce the number of errors that occur while solving the puzzle. It is interesting to note that the smaller LLMs have consistently low performance with the increase in the grid size of the puzzles but the larger LLMs have mixed performance. 

\subsection{Reasoning Chain Evaluation}


\begin{figure}
    \centering
    \includegraphics[width=7.5cm]{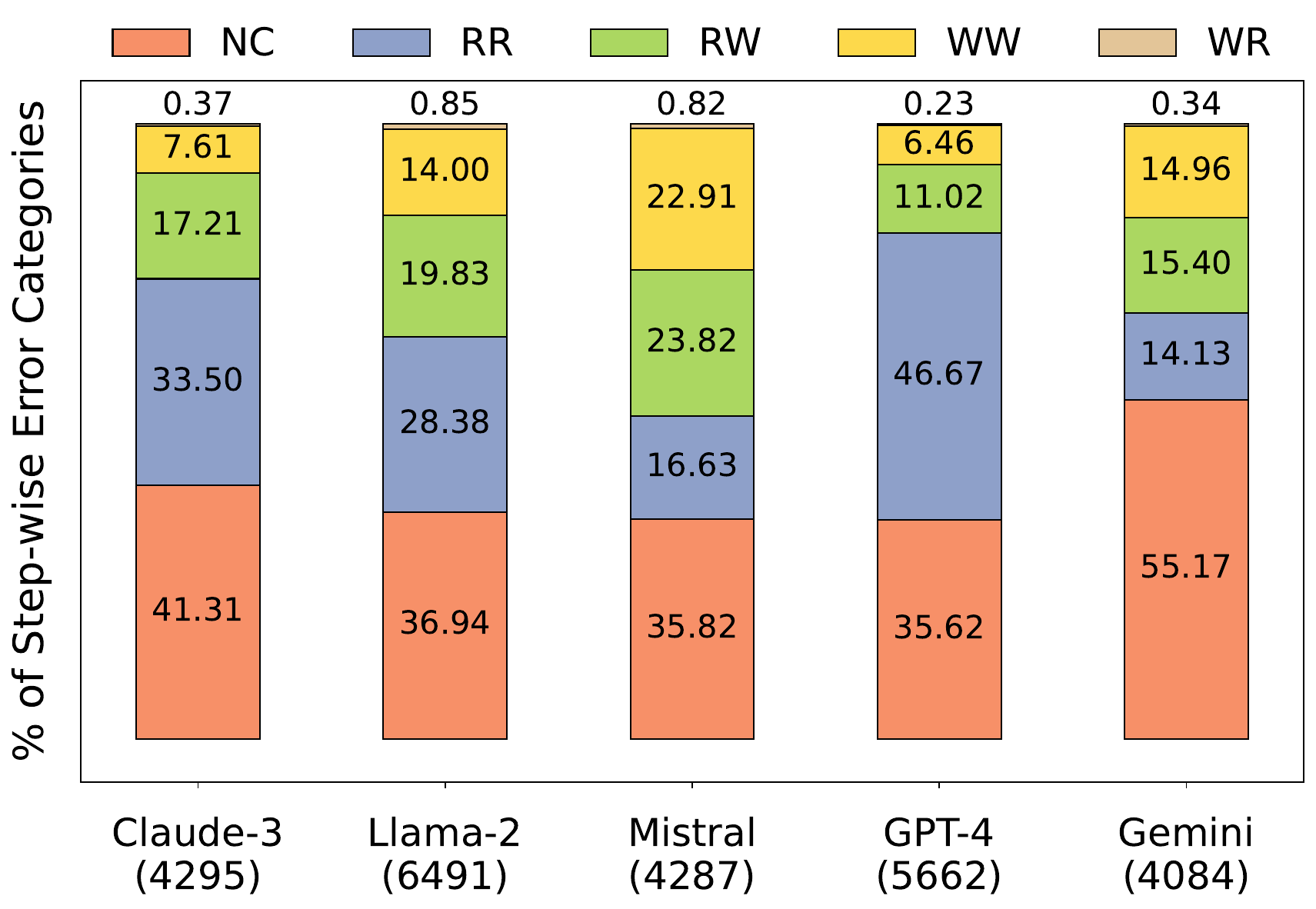}
    \caption{The percentage distribution of the broad error categories across the combined reasoning steps of all five LLMs. The total number of steps generated by each model is provided inside the round brackets below the model names.}
    \label{fig:broad-cat}
\end{figure} 

\begin{figure*}
    \centering
    
    \begin{subfigure}[t]{0.32\textwidth}
        \centering
        \includegraphics[width=\textwidth]{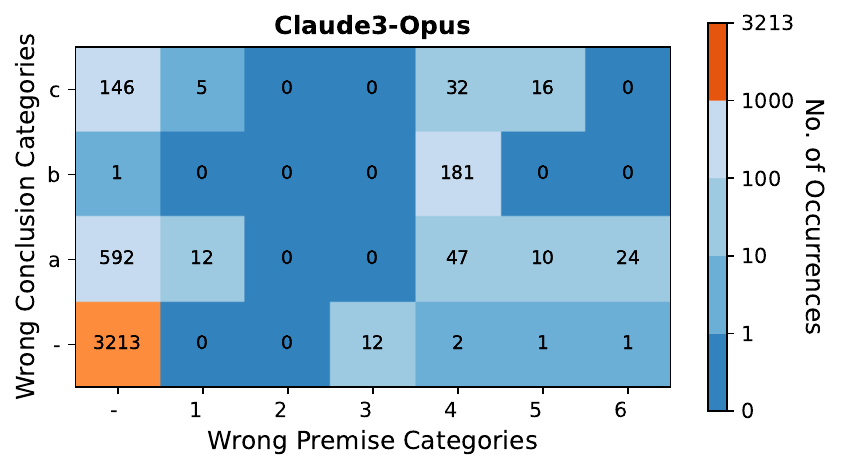}
    \end{subfigure}
    \hfill 
    \begin{subfigure}[t]{0.32\textwidth}
        \centering
        \includegraphics[width=\textwidth]{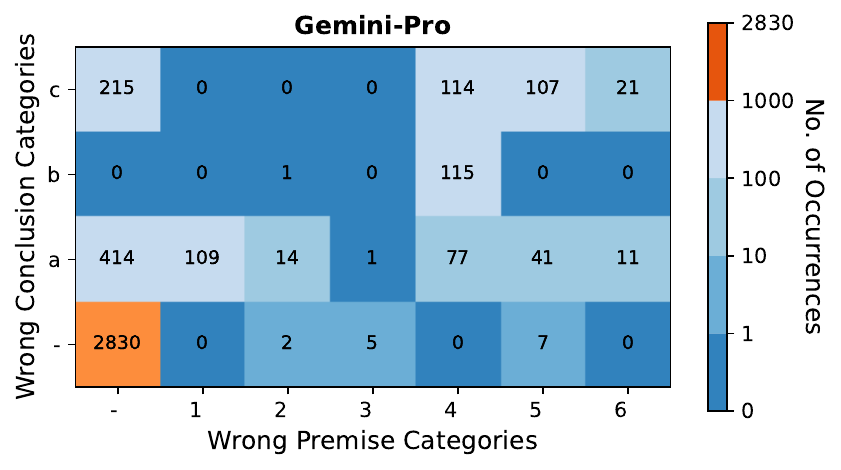}
    \end{subfigure}
    \hfill
    \begin{subfigure}[t]{0.32\textwidth}
        \centering
        \includegraphics[width=\textwidth]{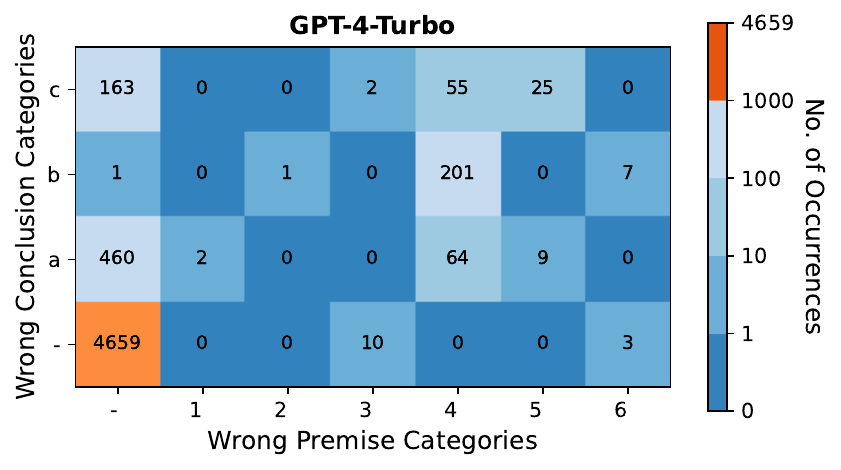}
    \end{subfigure}
    
    \vspace{1em} 

    \begin{subfigure}[t]{0.32\textwidth}
        \centering
        \includegraphics[width=\textwidth]{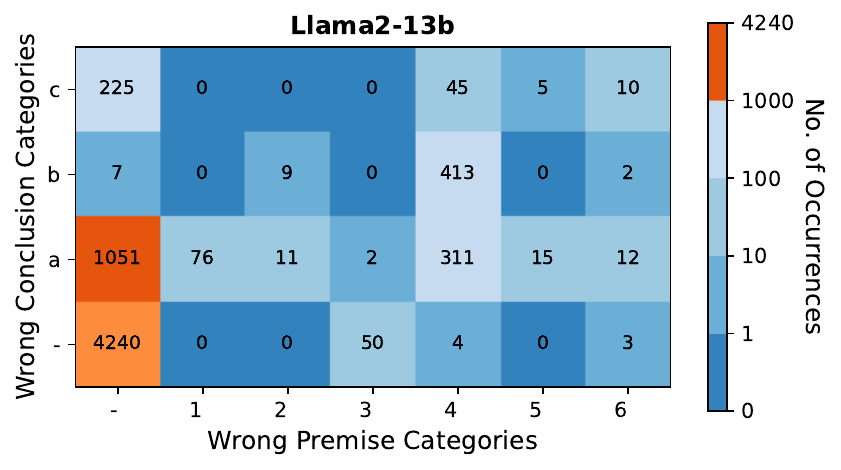}
    \end{subfigure}
    \hfill
    \begin{subfigure}[t]{0.32\textwidth}
        \centering
        \includegraphics[width=\textwidth]{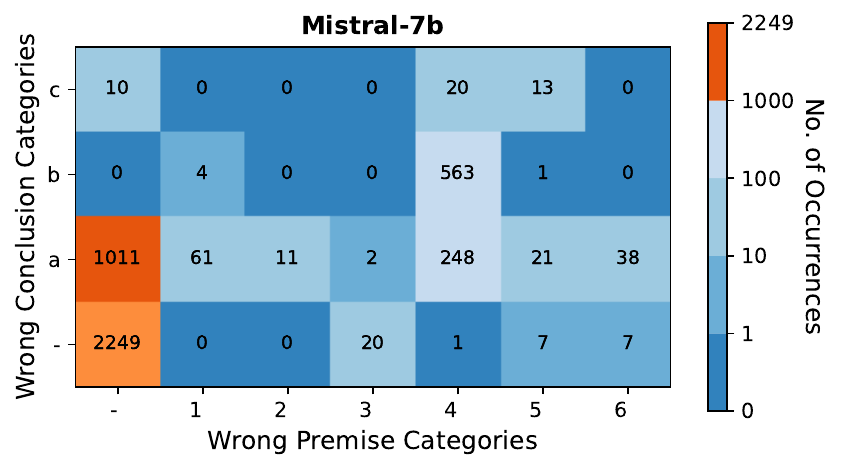}
    \end{subfigure}
    \hfill
    \begin{subfigure}[t]{0.32\textwidth}
        \centering
        \includegraphics[width=\textwidth]{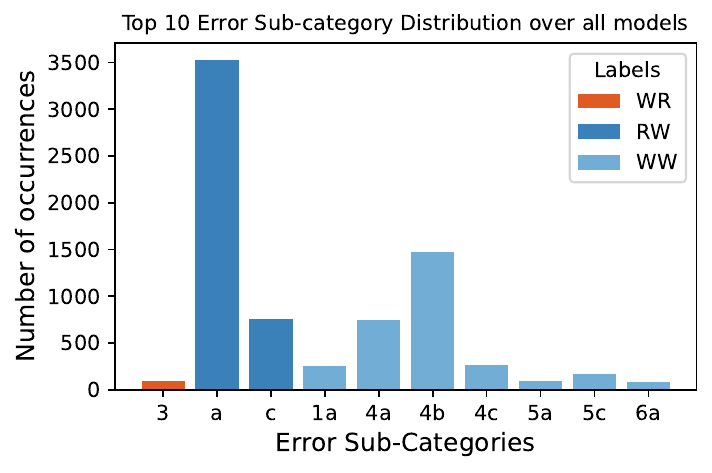}
    \end{subfigure}

    \caption{The first five sub-figures in the above section show the error Sub-category distribution over five LLMS. The last sub-figure denotes the top 10 error Sub category distribution across all model reasoning steps.}
    \label{fig:autoeval}
\end{figure*}

The relative distribution of the broad error categories over the collective reasoning steps for each model is given in Figure \ref{fig:broad-cat}. It is important to note that, despite using the same zero-shot-CoT setting, the GPT-4 and Llama-2 used significantly more reasoning steps ($> 5.5k$ steps) to solve the 274 puzzles compared to the other three models ($\sim4k$ steps). The distribution of error sub-categories for each model is presented as heatmaps in the first five sub-figures in Figure \ref{fig:autoeval}. Here, we present several findings based on the evaluation of different error category distributions across \textit{GridPuzzle}.

\paragraph{Majority of reasoning steps are error-free.} 

Figure \ref{fig:broad-cat} shows that most reasoning steps for each model fall into the ``NC'' error category, indicating that many steps reiterate the facts or clues from the initial puzzle rather than focusing on reasoning. Over 55\% of Gemini-Pro's reasoning steps fall into this category, the most among all models, suggesting that Gemini spends the fewest steps on actual reasoning. The "RR" category comprises over 46\% of GPT-4's reasoning steps, highlighting its strong reasoning ability. This higher number of correct reasoning steps correlates with GPT-4's higher \textit{PuzzleEval} score, reflecting its overall effectiveness.

\paragraph{The accuracy is low despite the reasoning chains being mostly error-free.}
The disparity between accuracy and \textit{PuzzleEval} arises from the relative location of errors within the reasoning chains. It has been observed that ``RR'' category reasoning steps mainly occur in the initial half of the chain, leading to a high overall \textit{PuzzleEval} score. Conversely, errors in the ``RW'', ``WR'', and ``WW'' categories typically occur in the latter half, resulting in incorrect final answers and lower accuracy scores. Based on error taxonomy, ``RW'', ``WR'', and ``WW'' broad error categories have been further dissected into $6\times3$ error sub-categories, with their distribution across reasoning steps shown in Figure \ref{fig:autoeval}.


\paragraph{Dominant broad categories: RW and WW.} The most common error sub-category across all heatmaps appears to be the ``-'', the absence of errors. All the reasoning steps with ``NC'' and ``RR'' classifications fall in this category. To observe the actual overall trend across all 5 LLMs, the top 10 most common error sub-categories have been listed in the last sub-figure of Figure \ref{fig:autoeval}. The top categories `a' and `c' refer to the \textbf{Wrong Reasoning} and the \textbf{Wrong Elimination} sub-categories under the ``RW'' category. These errors arise when the premise is correct but the LLMs fail to make accurate deductions from it. A number of the top 10 sub-error categories (`1a', `4a', `4b', `4c', `5a', `5c', and `6a') emerge from the ``WW'' category. 

For the categories, `4a', `4b', and, `4c' the errors in the premise are propagated from errors in previous reasoning steps showing how initially occurring errors in the chain can lead to more dependent errors. The `4b' error category is the one where this behavior is maximized as here both the premise and conclusions were wrong because of previously propagated errors. The `5a' and `5c' errors occurred due to the incompleteness or lack of information in the premise and wrong reasoning or elimination in the corresponding conclusions. The `1a' kind of error occurred when the premise consisted of hallucinated information. The only sub-category from the ``WR'' category making it in the top 10 is the `3' category which is caused due to wrong assumptions in the premise.  It can be noted here that the reasoning steps of the ``WR'' category do not deteriorate either of the evaluation metrics, as the conclusions ended up being correct, but rather indicate the inconsistency of the LLMs in reasoning over puzzle-solving.

\begin{table}[!htbp]
\small
\centering
\begin{tabular}{lcc}
\toprule
Mitigation Strategy   & Accuracy  & \textit{PuzzleEval}\\ 
\midrule
Baseline   & 12  & 0.61 \\ 
Plan-and-Solve   & 9  & 0.62 \\ 
Self-correct   & 10  & 0.59 \\ 
Self-discover   & \textbf{13}  & \textbf{0.65} \\ 
Feedback-Learning   & 10  & 0.59 \\
Program-of-Thought   & 10  & - \\
\bottomrule
\end{tabular}
\caption{The results for accuracy and \textit{PuzzleEval} using GPT-4-Turbo, with and without mitigation strategies for the 60 samples of $3 \times 4$ grid-size.}
\label{tab:mitstrat}
\end{table}

\paragraph{Proprietary LLMs are way better at GridPuzzle than Open-Source LLMs.}
From the results of objective and subjective metrics, it is evident that the open-source models have lower performance on the grid-puzzle-solving task than the proprietary models. The Llama-2 and Mistral models have the lowest accuracy values and their low performance on the \textit{PuzzleEval} consistently degrades with the increase in the size and complexity of the grids. The Claude-3, Gemini, and GPT-4 models have higher values of accuracy but their performance across the grid sizes in the \textit{PuzzleEval} is inconsistent. The disparity in the performance of both kinds of models can be attributed to the difference in their parameter sizes and the low instruction following capabilities of small open-source models.

\paragraph{Popular reasoning error mitigation strategies do not improve LLMs on GridPuzzle.} We conduct a case study on a subset of \textit{GridPuzzle}, focusing on a 3x4 grid size, utilizing commonly employed prompting techniques to enhance the reasoning capabilities of LLMs. In particular, we use five strategies: (1) Plan-and-Solve \cite{Wang2023PlanandSolvePI}, (2) Self-correct \cite{zhang2024small}, (3) Self-discover \cite{zhou2024selfdiscover}, (4) Feedback-Learning, and (5) Program of Thought prompting \cite{chen2023programthoughtspromptingdisentangling}. We updated the prompts corresponding to these techniques to include some of our major findings from the reasoning chain evaluations and error categorization analysis as precautionary instructions. 

The first strategy Plan-and-Solve prompts the model to first generate a plan to solve the given problem and then follow those steps. The second strategy is inspired by the Self-correct method which uses a combination of self-verification and self-refine to improve reasoning. Next, we used the Self-discover technique which is a 2-step structured reasoning process. We created our prompting technique called ``Feedback-Learning'' by providing specific feedback system instructions to the LLM based on our error taxonomy. Lastly, we also implemented a code-style prompting technique that implements a code to solve the puzzle but does not give a reasoning chain. The detailed prompt structure is described in Appendix \ref{Appendix:mitstrat} and the results of these strategies are in Table \ref{tab:mitstrat}. It is evident from the results that prompting-based strategies are not sufficient to significantly improve the LLM reasoning on the grid-puzzle-solving task. Compared to the rest of the strategies, Self-Discover marginally improves the performance on both accuracy and \textit{PuzzleEval}. These results indicate the need to develop techniques beyond prompting by having deeper insights from LLMs' reasoning chains.

\section{Conclusion}

In this work, we evaluated the logical reasoning abilities of LLMs through the lens of a grid-based puzzle-solving task. We introduced \textit{GridPuzzle}, an evaluation dataset of 274 puzzles with various grid sizes. From a manual evaluation of reasoning chains generated by five different LLMs on \textit{GridPuzzle}, we developed a fine-grained error taxonomy with five broad categories and nine sub-categories. We then created an Auto-evaluator to automate the identification of error categories, providing broader insights into error distributions across the dataset. Additionally, we proposed \textit{PuzzleEval}, a reference-free metric to objectively evaluate the correctness of reasoning chains for grid-based puzzles. Our analysis of error distributions in \textit{GridPuzzle} revealed several interesting findings and insights into the logical reasoning abilities of different LLMs. We further evaluated existing reasoning-specific prompting methods, such as self-discover and self-correct, finding that they do not improve results on \textit{GridPuzzle}. We believe our work offers a challenging dataset, highlights where these LLMs make mistakes, and provides insights to develop better logical reasoning systems for complex tasks such as grid puzzle-solving.

\section*{Limitations}
While \textit{GridPuzzle} facilitates the evaluation of LLMs' logical reasoning abilities, the complexity of the puzzles can be enhanced by incorporating further complex grid sizes beyond 4x6. Additionally, this study can be extended to different types of puzzles, such as Sudoku, Game of 24, and commonsense puzzles. Though our study provides fine-grained error categories, it can be further refined by mapping to formal logic to identify more detailed and atomic errors, offering a deeper understanding of LLMs' reasoning failures. Although we propose an effective automatic method for error identification to reduce manual analysis, exploring other automated methods using smaller-scale supervised learning could be a promising future research direction. We also note that this research is currently limited to the English language and can be extended to multilingual scenarios to evaluate the logical reasoning abilities of LLMs.


\section*{Ethics Statement}

The dataset, GridPuzzle, used for this study is based on 274 puzzles from the open-source platform (more details in section 3.1). No personal information from data creators has been collected during the creation of the dataset. The data collection process strictly adheres to the terms of use and privacy policies of the platform. Furthermore, the use of proprietary LLMs such as GPT-4, Gemini, and Claude-3 in this study adheres to their policies of usage. We have used AI assistants (Grammarly and ChatGPT) to address the grammatical errors and rephrase the sentences.

\section*{Acknowledgement}

We thank the anonymous reviewers for their constructive suggestions. We extend our gratitude to the Research Computing (RC), and Enterprise Technology at ASU for providing computing resources, and access to the ChatGPT enterprise version for experiments. We acknowledge support by a 2023 Spring Amazon Research Award (ARA). This material is also based upon work supported by the Engineering Research and Development Center - Information Technology Laboratory (ERDC-ITL) under Contract No. W912HZ24C0022.

\bibliography{custom}

\clearpage

\appendix
\section{\textit{GridPuzzle} Dataset - Sample Puzzle}
\label{appendix:puzzlesample}
\begin{figure}[ht]
    \centering
    \includegraphics[width=7.5cm]{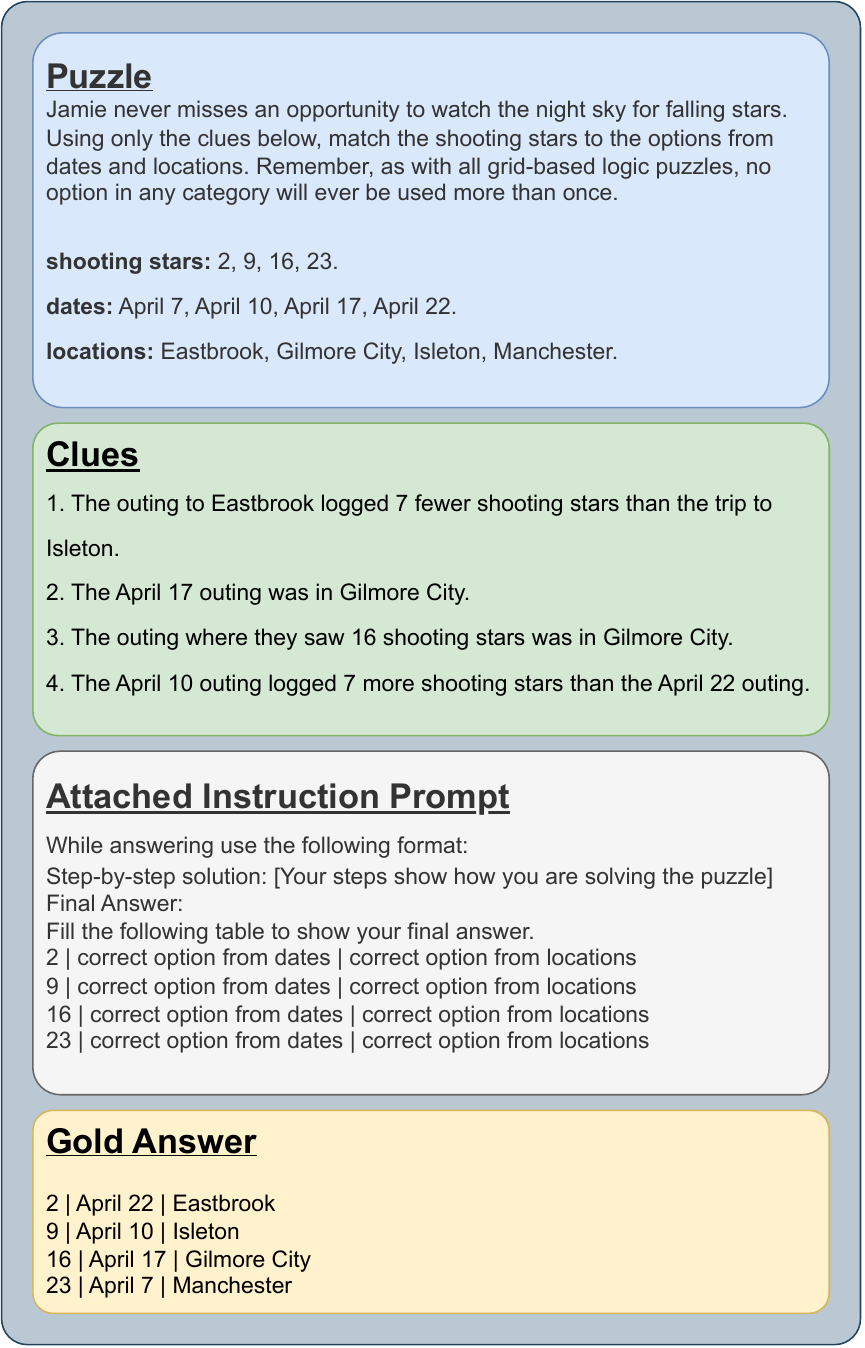}
    \caption{The prompt structure of a 4 x 4 grid size puzzle from \textit{GridPuzzle} dataset. Every Zero-shot-CoT prompt from the dataset consists of Puzzle, its corresponding Clues, the Instruction for solving the puzzle, along with the Gold solution of the Puzzle.}
    \label{fig:cotprompt}
\end{figure}

The \textit{GridPuzzle} dataset contains 274 puzzles of various grid sizes and complexity. A sample puzzle from the dataset along with the Zero-shot-CoT prompt is described in Figure \ref{fig:cotprompt}. All the puzzles in the dataset have a similar structure with varying numbers of clues.

\section{Auto Evaluator: GPT-4o}
\label{appendix:autoevaluator}
\begin{figure}[ht]
    \centering
    \includegraphics[width=0.98\linewidth]{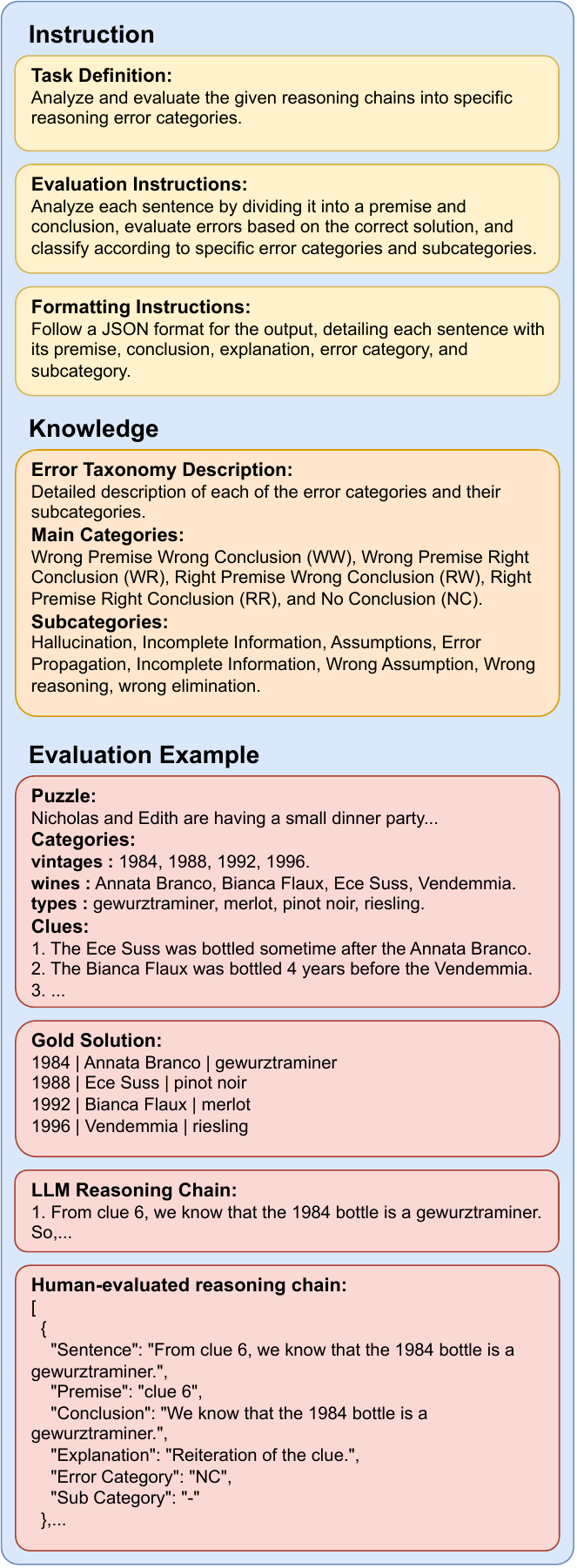}
    \caption{The prompt structure of auto-evaluator. The prompt consists of instructions to evaluate reasoning chains, Knowledge of the error category descriptions, and a human-evaluated reasoning chain example. This system prompt is followed by a user prompt with the puzzle, its gold solution, and the reasoning chain to be evaluated.}
    \label{fig:autoeval_prompt}
\end{figure}

To expand the reasoning chain evaluation process we prompt the GPT-4o model with a detailed system prompt. The structure of this system prompt is elaborated in Figure \ref{fig:autoeval_prompt}. The 3 main components of this system prompt are the \textbf{Instruction} - similar to the ones given to human evaluators, the \textbf{Knowledge} - obtained from the error taxonomy, and an \textbf{Exemplar} - consisting of a Puzzle, its Gold Solution, the LLM-generated Reasoning chain, and the evaluated Reasoning Chain.

\clearpage

\begin{figure*}[ht]
    \centering
    \includegraphics[width=0.75\linewidth]{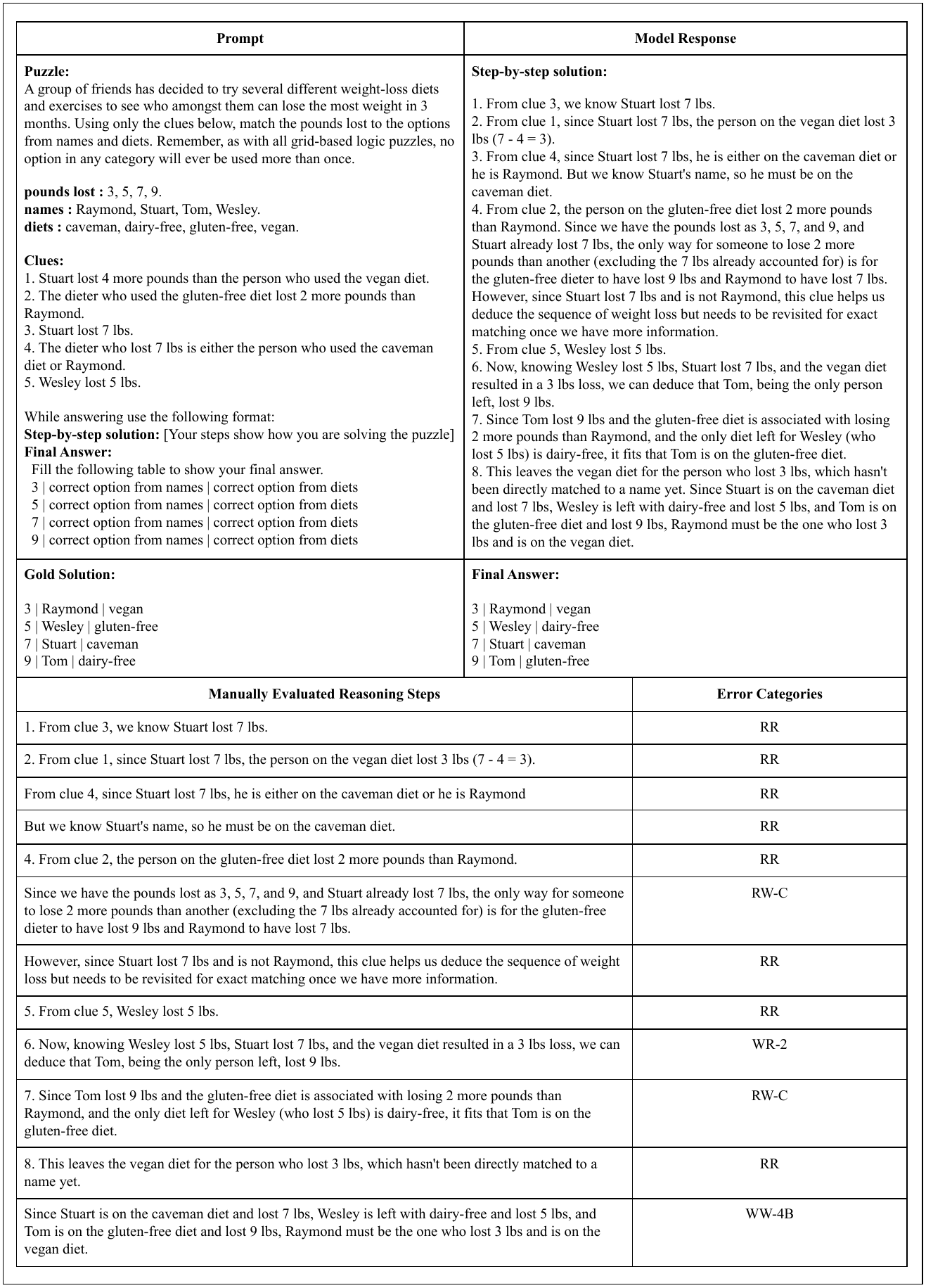}
    \caption{The \textbf{top left} section of the figure consists of a 3x4 sample puzzle from the \textit{GridPuzzle} dataset along with the Zero-shot-CoT prompt. Right below the prompt, we have the Gold solution for the corresponding puzzle. In the \textbf{top right} section of the figure, we have the Model-generated Reasoning chain to solve this puzzle along with the Final Answer. In this particular instance, the reasoning chain was generated by the GPT-4 model. In the \textbf{bottom half} of the figure, we have the manually evaluated reasoning chain, split into individual sentences. The corresponding error categories found in each step is given the \textbf{bottom right} column called `Error Categories.'}
    \label{fig:manual_annotation}
\end{figure*}

\begin{figure*}[ht]
    \centering
    \includegraphics[width=0.75\linewidth]{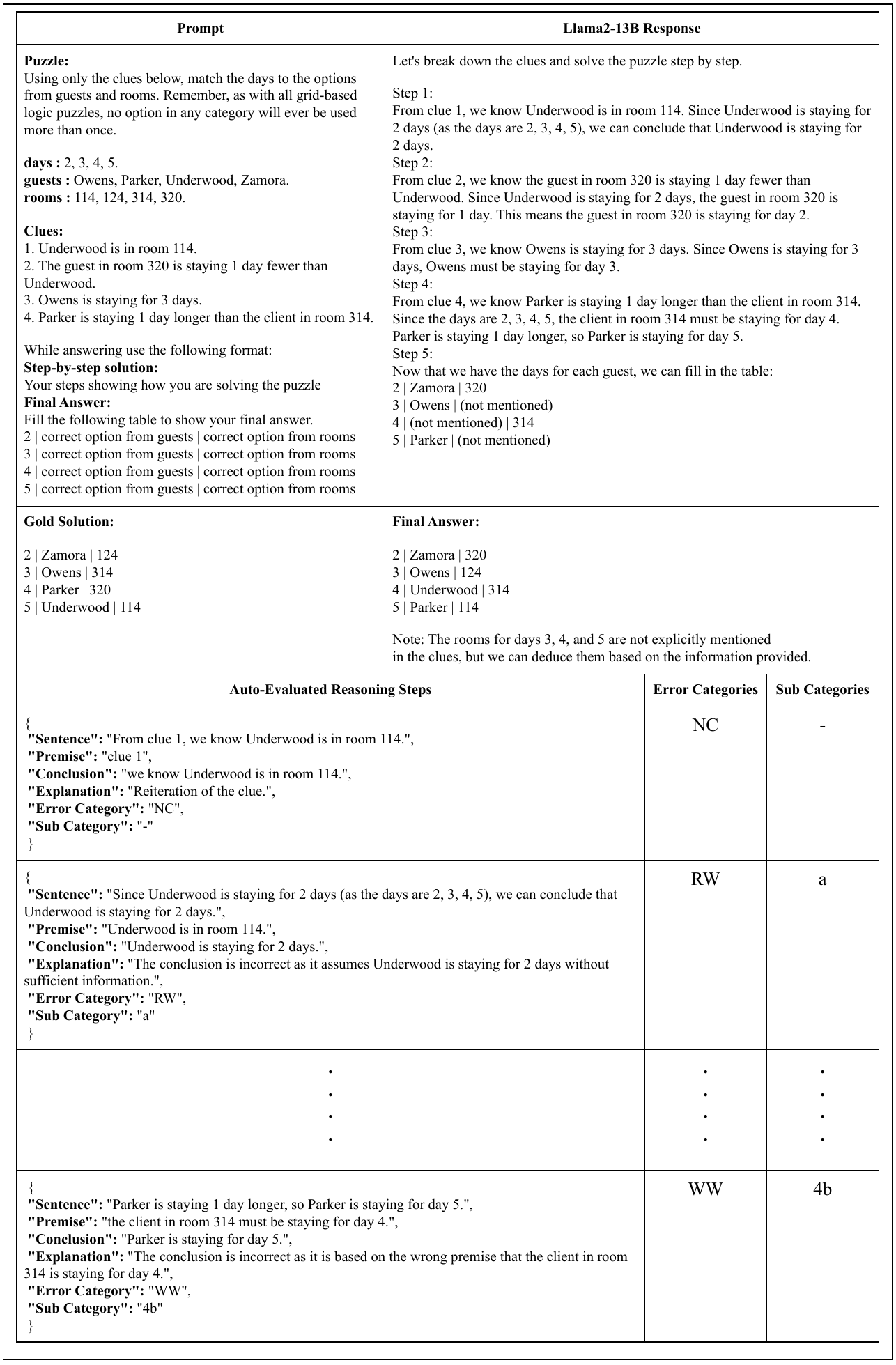}
    \caption{The \textbf{top left} section of the figure consists of a 3x4 sample puzzle from the \textit{GridPuzzle} dataset along with the Zero-shot-CoT prompt. Right below the prompt, we have the Gold solution for the corresponding puzzle. In the \textbf{top right} section of the figure, we have the Model-generated Reasoning chain to solve this puzzle along with the Final Answer. In this instance, the reasoning chain was generated by the Llama2-13b model. In the \textbf{bottom half} of the figure, we have the GPT-4o Auto-Evaluated Reasoning chain.The auto-evaluation is done sentence-wise and the output is in a JSON-structured format consisting of 5 components: the \textit{Sentence}, the \textit{Premise}, the \textit{Conclusion}, the \textit{Error category} and the \textit{Sub-category}. The corresponding error categories found in each sentence are given in the \textbf{bottom right} columns called `Error Categories' and `Sub Categories.'}
    \label{fig:auto_annotation}
\end{figure*}

\clearpage

\section{Evaluation of Reasoning Chains}
\label{Appendix:rcevaluation}
In order to identify the error categories from the erroneous reasoning chains we conducted manual and auto-evaluation of the reasoning chains. The process of manual evaluation has been described in figure \ref{fig:manual_annotation} and the process of auto-evaluation using GPT-4o has been described in Figure \ref{fig:auto_annotation}.

\section{Annotation Guideline}
\label{Appendix:guideline}
\begin{figure}[ht]
    \centering
    \includegraphics[width=0.95\linewidth]{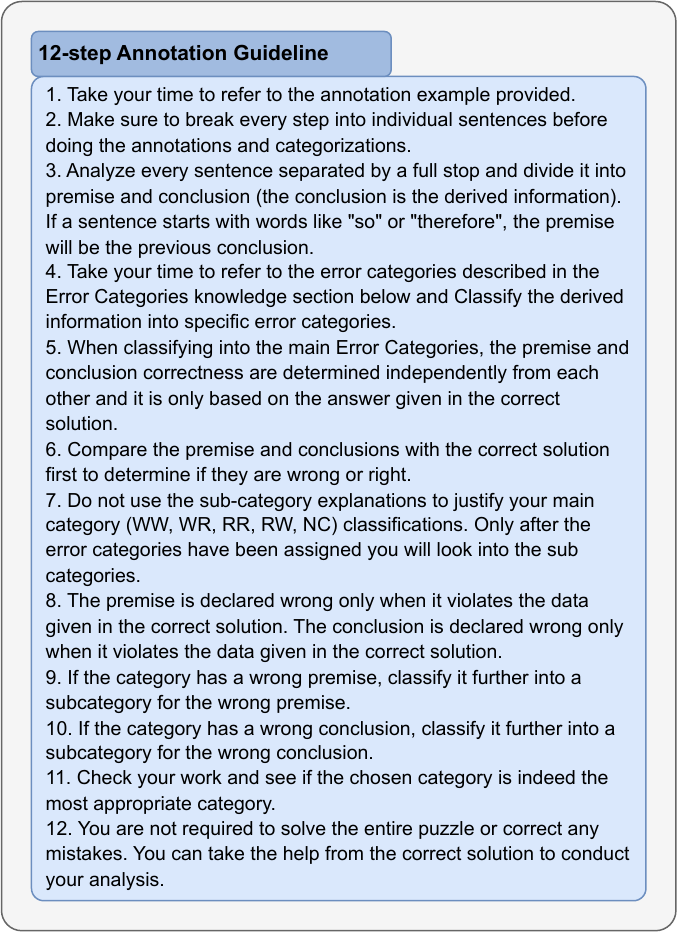}
    \caption{The 12-step guideline provided to the annotators to conduct manual evaluation of the reasoning chains.}
    \label{fig:anot_guideline}
\end{figure}

To conduct the manual analysis of the reasoning chains, the annotators were provided the guidelines described in figure \ref{fig:anot_guideline}. The same guideline was also used to create the system prompt for the GPT-4o Auto-evaluator. The annotation process was conducted by 5 annotators and the annotations were also cross-examined to resolve any discrepancies. 

\section{Mitigation Strategy Prompts}
\label{Appendix:mitstrat}
\begin{figure}[ht]
    \centering
    \includegraphics[width=0.95\linewidth]{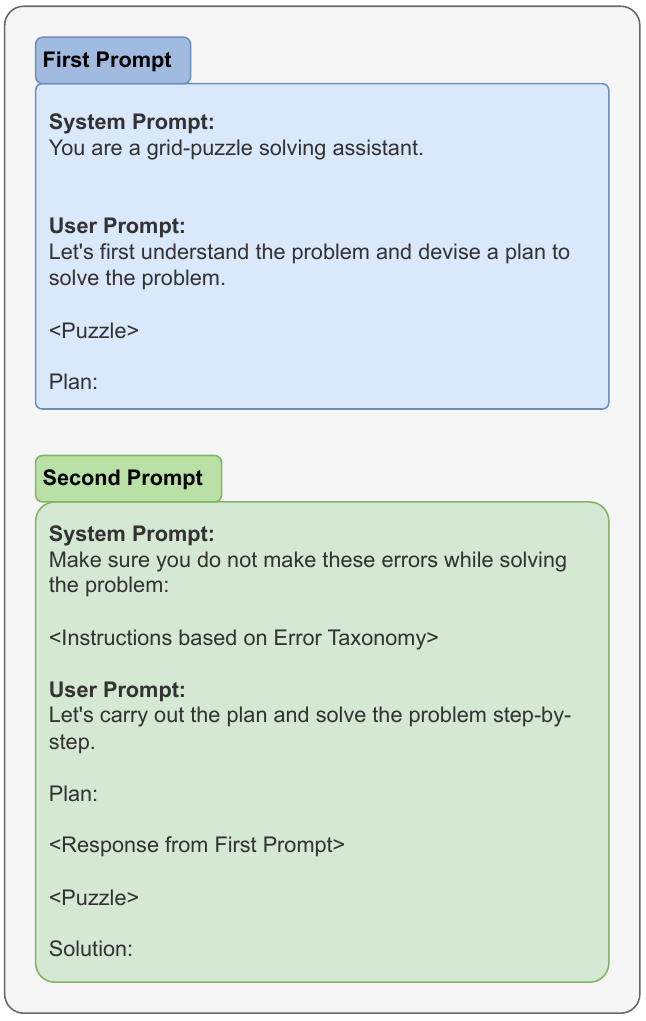}
    \caption{The prompt structure for the Plan-and-Solve strategy which is split into two prompts one for planning and the other for solving the puzzle.}
    \label{fig:planandsolve}
\end{figure}

We conducted a study on the 60, 3x4 puzzles present in \textit{GridPuzzle} dataset to try and improve the reasoning abilities of LLMs when solving the grid-puzzle task. We used prompt-based methods, such as the Plan-and-Solve technique, which divides puzzle-solving into planning and solving steps. We also enhanced the solver with insights from our error taxonomy. The prompt structure for this technique is given in figure \ref{fig:planandsolve}. 

\begin{figure}[ht]
    \centering
    \includegraphics[width=0.95\linewidth]{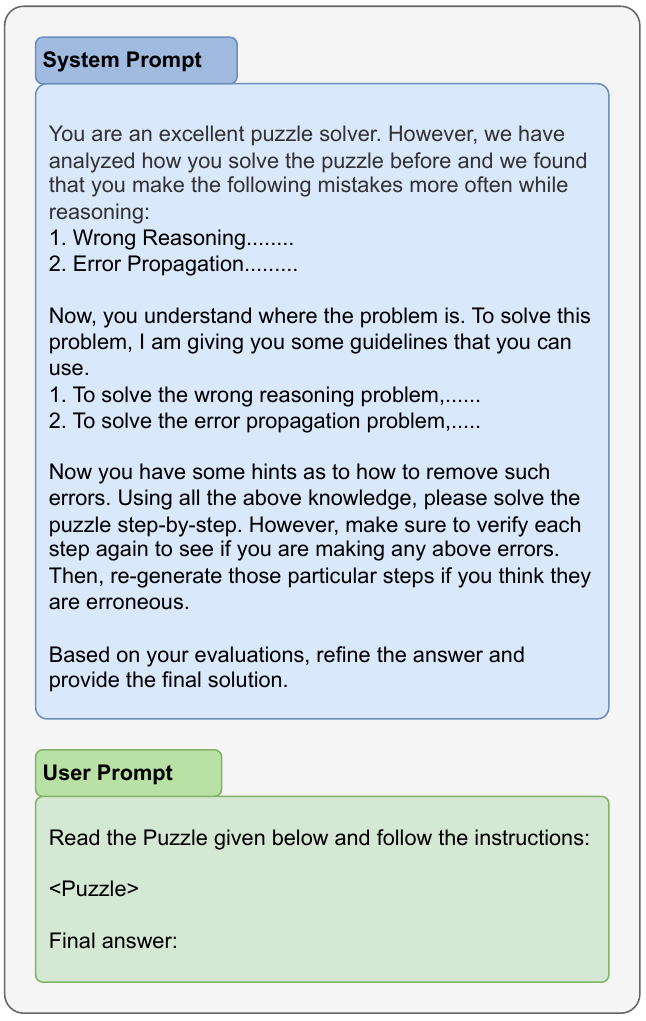}
    \caption{The prompt structure for the Feedback-learning strategy. The system prompts consist of instructions regarding the major errors as well as ways to rectify those errors.}
    \label{fig:feedback}
\end{figure}

Next, we devised our own strategy to improve LLM reasoning by using the top error categories from our findings and teaching the LLM to rectify those mistakes. This strategy termed as Feedback-learning makes use of a detailed system prompt that acts as a feedback-providing unit followed by a basic user prompt to solve the puzzle. The prompt structure for this strategy is shown in figure \ref{fig:feedback}.

\begin{figure}[ht]
    \centering
    \includegraphics[width=0.95\linewidth]{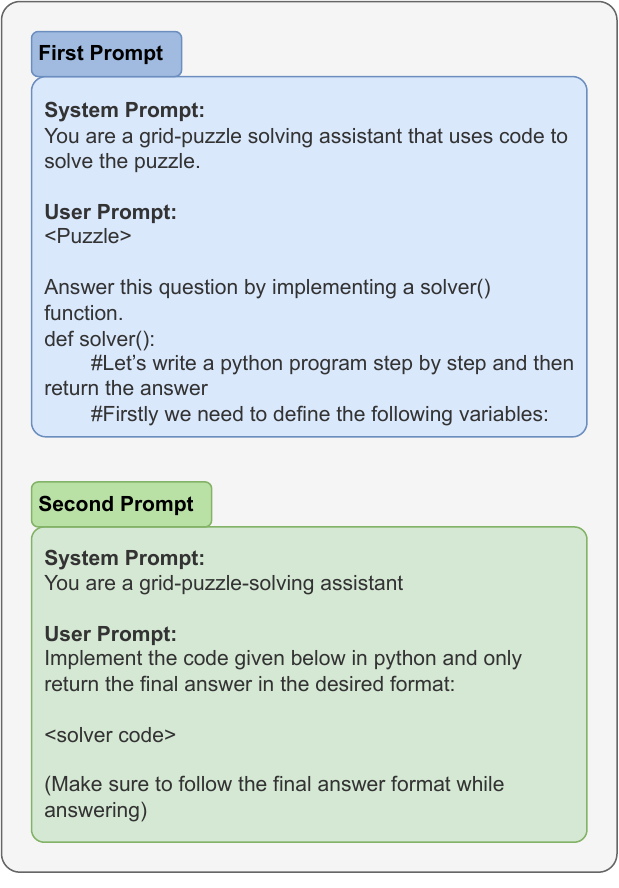}
    \caption{The prompt structure for the Program of Thought prompting technique. In the first part, we asked the LLM to generate a code to solve the given puzzle. In the second part, an LLM is prompted to implement the code produced in the first part to get the Final Answer table.}
    \label{fig:pot}
\end{figure} 

\begin{figure}[ht]
    \centering
    \includegraphics[width=0.95\linewidth]{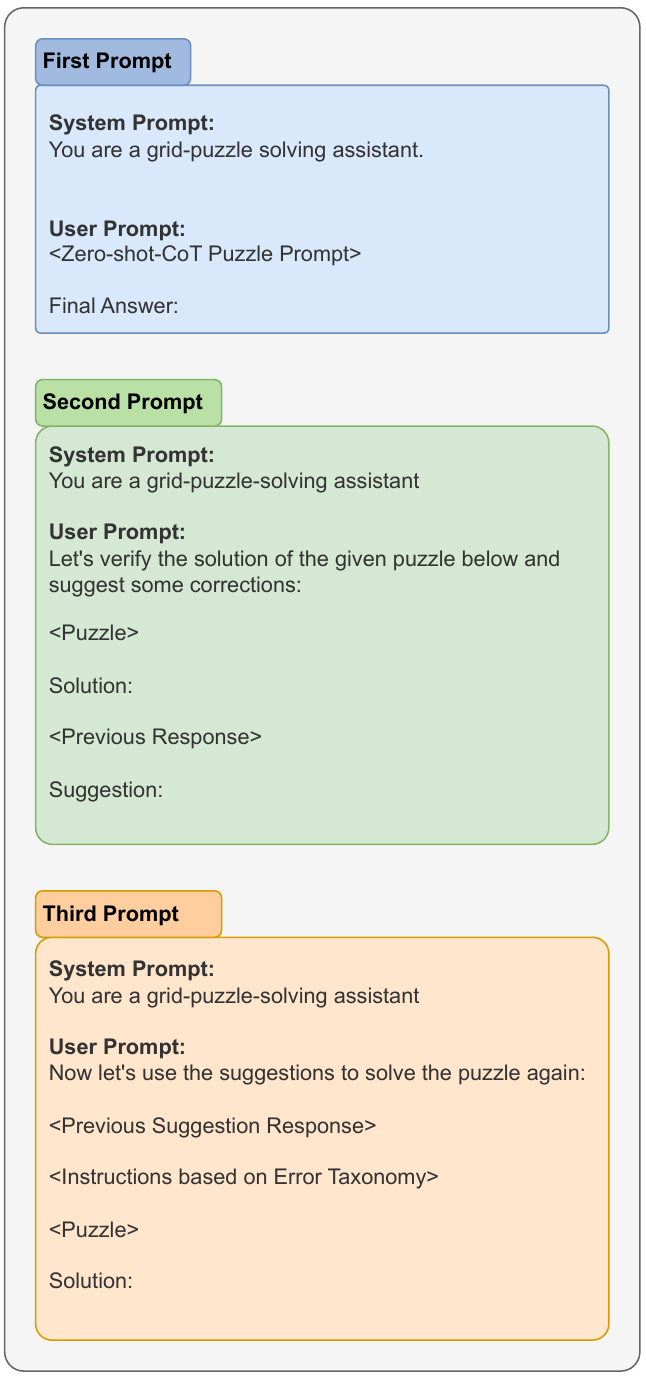}
    \caption{The prompt structure for the Self-Correct strategy is split into 3 parts. The first prompt solves the puzzle, the second prompt verifies the solution and gives suggestions to improve the solution, and the third prompt uses these suggestions along with error taxonomy-based instructions to refine the final solution. }
    \label{fig:selfcorrect}
\end{figure}

\begin{figure}[ht]
    \centering
    \includegraphics[width=0.95\linewidth]{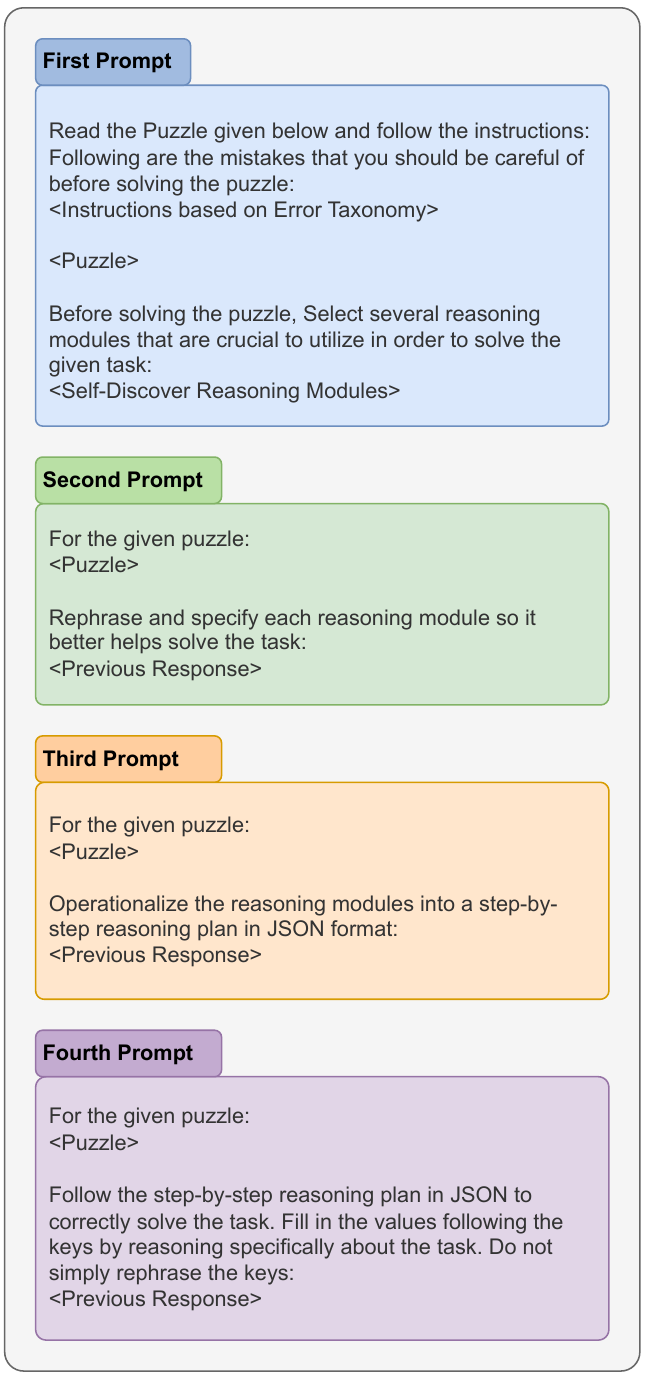}
    \caption{The prompt structure for the Self-Discover strategy. In the first part of this prompt the model is prompted to assess the problem and select the appropriate reasoning module to solve it. Then the module is modified to give a structured plan to solve the puzzle. In the second part, the model uses this structured plan along with instructions from our error taxonomy to solve the puzzle.}
    \label{fig:selfdiscover}
\end{figure}

 We also implemented a code-based technique to sole GridPuzzle. We borrowed the PoT prompt from the original implementation to create a solver function to solve the puzzles. Next, we asked an LLM to implement this code and produce the Final Answer. Since the codes produced by the LLM may contain some errors we utilized the LLM's compiler to implement the code instead of a rigid Python compiler. The prompt structure is provided in figure \ref{fig:pot}. Next is the Self-correct strategy which merges Self-verify and Self-refine qualities to minimize LLM reasoning errors. It starts with solving the puzzle using a Zero-shot-CoT prompt, followed by prompting the LLM to verify and refine the solution. Finally, it integrates the model's suggestions with insights from our error taxonomy to enhance the puzzle-solving response. The prompt structure for this strategy is shown in figure \ref{fig:selfcorrect}. Lastly, the Self-Discover strategy, depicted in figure \ref{fig:selfdiscover}, proved most effective in reducing LLM reasoning errors in puzzle-solving. This approach begins by having the model analyze the problem and potential errors, follows with a list of prescribed reasoning modules, prompts the LLM to select and apply the most suitable module, and concludes by using a structured prompt to solve the puzzle.

\section{Model Scaling Effect: Llama-70B}
\label{Appendix:llama70b}

We conducted a case study on the Llama models to analyze their performance on GridPuzzle with increasing model parameter size. We repeated the same experiment in the Zeo-shot-CoT setting with the Llama-70B model. We found that the performance of the bigger model was marginally higher than the 13B model. The Accuracy went up from 1 correct final answer in the 13B model to 2 in the 70B model. The scores on PuzzleEval also went up 11\% on average. However, despite the slight improvement, the Llama model's performance was still inferior to GPT-4, Gemini, and Claude. The experimental findings are presented in Table \ref{tab:scaling}. We infer that even with the increasing model parameter size, the LLMs lack the intrinsic reasoning capabilities required to solve complex logic problems such as \emph{GridPuzzle}.

\begin{table}
\small
\begin{center}
\resizebox{\linewidth}{!}{
\begin{tabular}{c|ccccc|c}
\toprule
\textbf{Model}   & \textbf{3 x 4} & \textbf{3 x 5} & \textbf{4 x 4} & \textbf{4 x 5} & \textbf{4 x 6}  & \textbf{Avg} \\ 
\midrule
Llama-70B  & 0.51  & 0.51  & 0.52  & 0.58  & 0.42  & \textbf{0.52} \\ 

\bottomrule
\end{tabular}
}
\end{center}
\caption{The results for \textit{PuzzleEval} on the different grid sizes available in \textit{GridPuzzle} dataset in terms of ACS for Llama-70B. The Accuracy of Llama-70B was 2/274 puzzles.}
\label{tab:scaling}
\end{table}

\end{document}